\documentclass[letterpaper]{article}
\usepackage{isea}
\usepackage[pdftex]{graphicx}
\usepackage{times}
\usepackage{amsmath}
\usepackage{amsfonts}
\usepackage{hyperref}
\usepackage{helvet}
\usepackage{courier}
\usepackage{url}
\usepackage[numbers]{natbib}
\title{Leveraging Generative AI Models to Explore Human Identity}
\author{Yunha Yeo\\
Graduate School of Culture Technology, KAIST\\
Daejeon, South Korea\\
yunhayeo@kaist.ac.kr\\
\newline
\newline
\And
Daeho Um\thanks{Corresponding author}
\\
Samsung Advanced Institute of Technology (SAIT)\\
Suwon, South Korea\\
daeho.um@samsung.com\\
}

\begin{document} 

\maketitle

\begin{abstract}
% 문장 1

This paper attempts to explore human identity by utilizing neural networks in an indirect manner. For this exploration, we adopt diffusion models, state-of-the-art AI generative models trained to create human face images. By relating the generated human face to human identity, we establish a correspondence between the face image generation process of the diffusion model and the process of human identity formation. Through experiments with the diffusion model, we observe that changes in its external input result in significant changes in the generated face image. Based on the correspondence, we indirectly confirm the dependence of human identity on external factors in the process of human identity formation. Furthermore, we introduce \textit{Fluidity of Human Identity}, a video artwork that expresses the fluid nature of human identity affected by varying external factors. The video is available at \url{https://www.behance.net/gallery/219958453/Fluidity-of-Human-Identity?}.

% By gradually adjusting the external input

% where the external input and the generated face image correspond to environmental aspects and resulting identity in the process of human identity formation.

% Building on previous scholars' suggestions for specific tools for examining the human mind, this paper proposes that AI can serve as a contemporary gateway to understanding human identity.
% This paper explores the fluidity of human identity through the application of AI diffusion models, drawing parallels between AI-generated image transformations and the evolving nature of the self.
% Specifically, we utilize the diffusion model, which gradually removes noise from images, to simulate how human identity develops in response to external influences.
% By manipulating paths in the color space and analyzing changes in output images, we reveal a new framework to discover characteristics of human identity.
% The novel interdisciplinary attempt to intertwine human identity and AI encourages further exploration at the intersection of AI and human identity.

% intro 보고 다시 볼게요
% -> Since it is impossible to thoroughly examine the human mind, which lies at the core of human nature, there have been studies investigating the human mind through the use of different tools, such as dreams and cameras.
% -> In this paper, we attempt to utilize artificial intelligence (AI) models for analyzing human identity.
% -> TBA

\end{abstract}


artificial intelligence (AI), AI-generated art, human psyche, human identity
% optical unconscious, diffusion model, AI-generated video, digital humanity, AI aesthetics, AI vision

\section{Introduction}

Humans have long sought to understand their nature, striving to explore the depths of the self. Beyond the physical aspects, the human psyche—encompassing the entire structure of the mind—has drawn significant attention from researchers. For instance, debates on human nature, which represent inherent characteristics, have persisted since ancient times, including discussions on whether humans are fundamentally good or evil. In this paper, among the many elements that comprise the human psyche, we focus on exploring human identity, a complex concept that defines who we are.

Despite the importance of the human psyche, research in this area faces critical challenges. The psyche is a highly complex system comprising various interconnected elements and relationships. Moreover, direct analyses involving humans are closely tied to ethical issues. To address this, there have been attempts to indirectly explore the human psyche using various tools as intermediaries. For instance, Sigmund Freud attempted to examine individuals' unconsciousness by analyzing dreams, proposing that dreams serve as a gateway to the human psyche.~\cite{freud1983interpretation} Photography, as insisted by Walter Benjamin, is another example of tools that let us analyze the concept of optical unconscious.~\cite{844f2d44-1ff5-349a-9fc9-5fcdcae52f99} While various tools, including dreams and photographs, have been proposed to explore the human psyche, interpretations based on such tools remain debatable due to the lack of concrete evidence connecting these superficial tools to the psyche.

Recently, rapid advancements in artificial intelligence (AI), particularly in neural networks that mimic the human nervous system, have ushered in an era of remarkable innovation.~\cite{jumper2021highly, krizhevsky2017imagenet} Neural networks are inspired by the structure and function of human neurons, replicating aspects such as information processing, learning mechanisms, and parallel processing. The use of neural networks in human psyche analyses has two key strengths. First, they share a clear structural connection with the human neural system. Second, neural networks can be utilized in experiments and simulations without ethical limitations.

\begin{figure}[t]
\includegraphics[width=\linewidth]{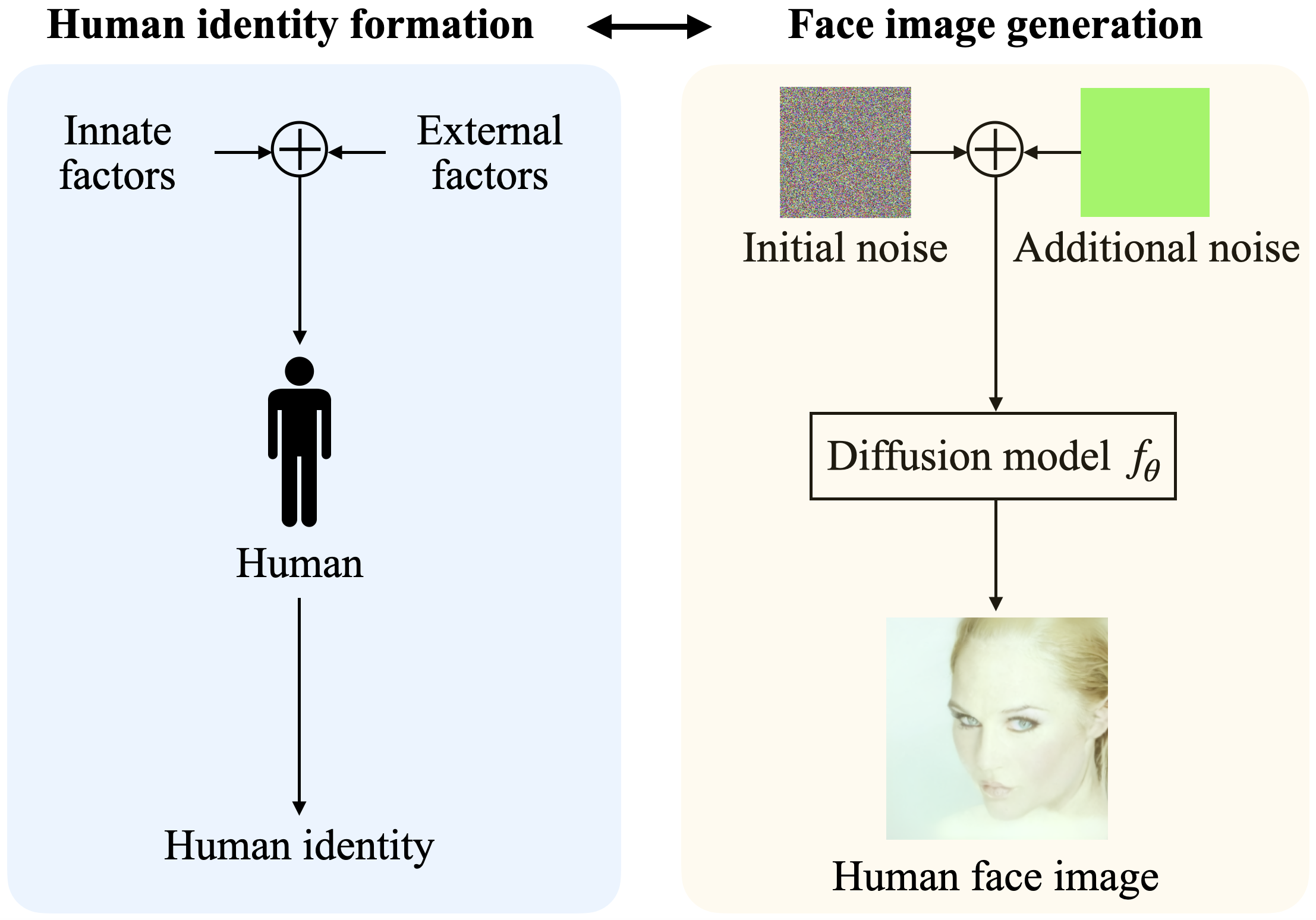}
\caption{Correspondence between the human identity formation process and the face image generation process.}
\vspace{-5mm}
\label{fig:teaser}
\end{figure}

\begin{figure*}[t]
\includegraphics[width=\linewidth]{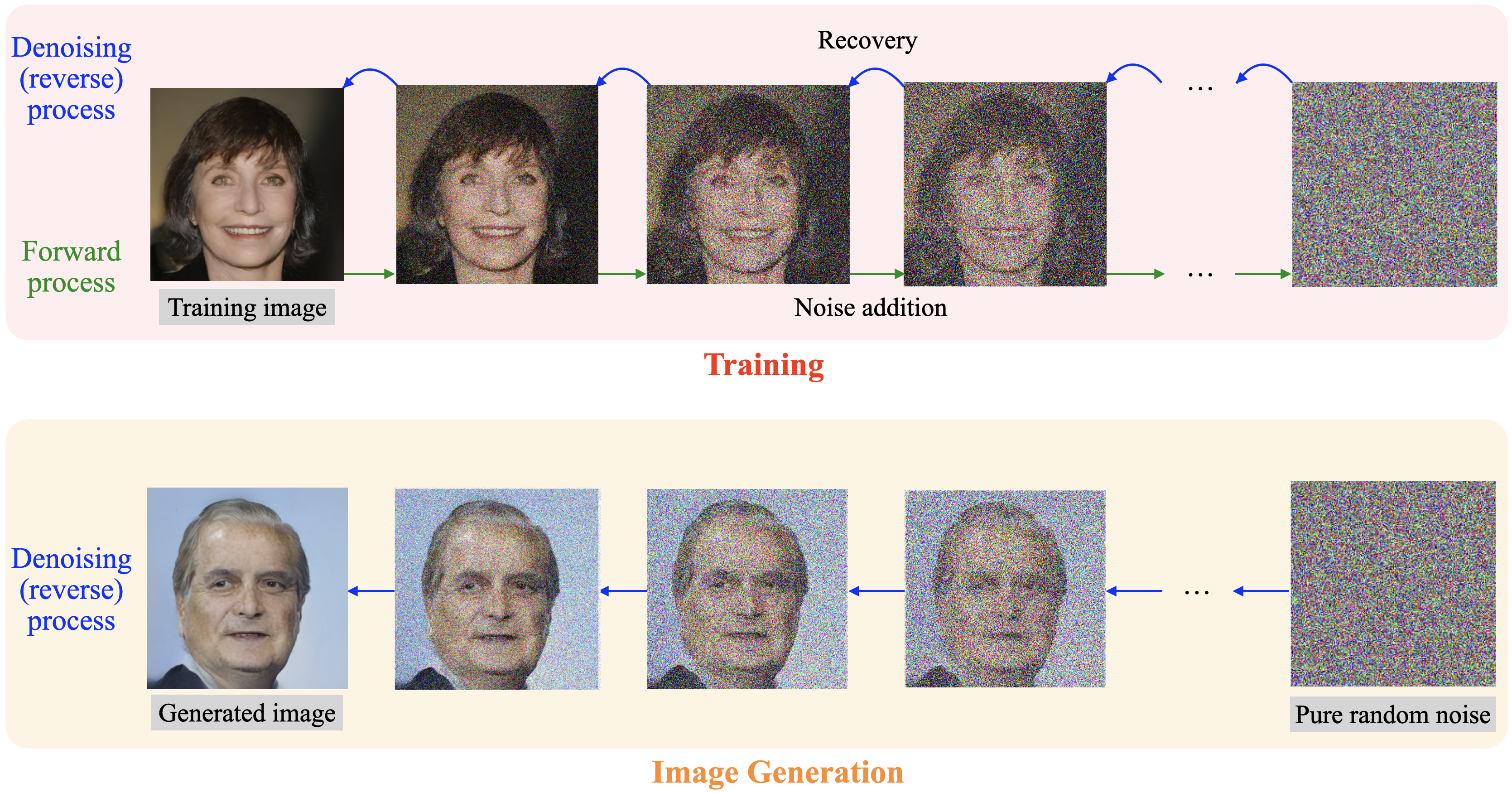}
\caption{The mechanism of diffusion models. The training stage consists of a forward process, where noise is incrementally added to a training image, and a denoising (reverse) process, where the model learns to progressively denoise and recover the image. After training, the diffusion model can generate a new image by applying the denoising process to pure random noise.}
\vspace{-3mm}
\label{fig:mechansim}
\end{figure*}

In this paper, we adopt neural networks, specifically diffusion models—state-of-the-art generative models—to analyze human identity.~\cite{yang2023diffusion, dhariwal2021diffusion} Diffusion models generate images from initial noise through a denoising process. In this process, we introduce additional noise into the initial noise. We utilize a diffusion model trained to generate human face images. Since human identity is known to be shaped by both innate and environmental aspects, we associate the initial noise with the innate aspects of identity formation. Similarly, we associate the additional noise with environmental influences provided by external input. Finally, we relate the human face generated by the diffusion model to human identity. In summary, we establish a correspondence between the neural network (\textit{i.e.}, the diffusion model) and human.

Here, we discover that when we inject additional noise of a specific color into the initial noise, the image generated by the diffusion model becomes imbued with that color. Based on this observation, we first fix the initial noise, which represents the innate factors that influence the formation of human identity. We then adjust the color of the additional noise to correspond to changes in external factors. By gradually modifying the color along various paths, we analyze the resulting changes in the generated human face image. We observe that as the color of the additional noise changes, the generated human face image changes continuously, not only in aspect of color but also in appearance. Through this causal relationship between the additional noise and the human face image, we indirectly investigate the causal relationship between external factors and human identity. Building upon this analysis, we create a video artwork \textit{Fluidity of Human Identity}  by sequentially connecting the continuously generated results, inviting viewers to engage in a deeper contemplation of human identity.

The main contributions of our work are three-fold:
\begin{itemize}
    \item We attempt an indirect analysis of human identity using neural networks, specifically diffusion models, by establishing a correspondence between the face image generation process of the diffusion model and the process of human identity formation.
    \item To the best of our knowledge, this is the first attempt to leverage neural networks to investigate the casual relationship between human identity and external factors, bridging the gap between neural networks and human identity studies.
    \item We develop \textit{Fluidity of Human Identity} representing the influence of external factors on human identity through the AI generative model. This artwork is not the result of human editing or manual manipulation but is entirely based on the algorithmic process within the AI generative model.
\end{itemize}

\section{Related Work}

Various philosophers and artists have explored the nature of human identity and its developmental processes. Within the framework of existentialist philosophy, Jean-Paul Sartre viewed the human self and identity as fluid, changing through the choices and actions individuals make. Sartre famously argued that ``existence precedes essence," meaning that humans do not possess a static essence but instead form their identity through their own decisions and experiences.~\cite{Sartre_Macomber_Elkaïm-Sartre_2007} Similarly, Gilles Deleuze and Félix Guattari’s concepts, such as Rhizome and Assemblage, also support the notion of human identity as an evolving, interconnected network rather than a static essence. Their ideas emphasize viewing the human self as part of a stream of constant reconstruction. ``In A Thousand Plateaus", they described identity as a nonlinear, multiple, and dynamic concept.~\cite{deleuze1987introduction} In summary, Sartre, Deleuze, and Guattari regarded human identity as a constantly evolving concept.

In the same vein, artists have attempted to visually express the fluid concept of human identity across a wide range of technologies, mediums, and periods. From the late 1970s to the 1980s, American photographer Cindy Sherman explored how one's appearance can be transformed depending on societal archetypes through her photography series \textit{Untitled Film Stills}. Sherman's work sparks discussions about the social construction of gender and identity, visually representing the idea that the self can change within the context of visual culture.~\cite{respini2012cindy}

Digital media artists have also explored themes of human identity. Mexican-Canadian electronic artist Rafael Lozano-Hemmer has created interactive installations that examine the intersection of technology, human behavior, and identity. His works often incorporate real-time data collection, including biometric information such as heartbeats and fingerprints, emphasizing the role of technology in shaping human identity and experience.~\cite{ref:hemmer} In particular, his installation \textit{Pulse Room} offers a profound exploration of the relationship between technology, art, and human identity. Using biometric data, the piece raises questions about what it means to be human in the digital age, specifically how technology mediates our perception of our bodies, emotions, and presence. Lozano-Hemmer's work serves as a metaphor for how we shape and contribute to collective identities in a technological world, providing a powerful reflection on human presence, individuality, and the interconnectedness of shared biological experiences.

Recently, artists have begun to utilize neural networks to generate art. Mario Klingemann, a German artist, explores visual imagination and the complex relationship between humans and technology by incorporating algorithms and AI.~\cite{ref:Klingemann_2018b} For example, in Klingemann’s notable artwork \textit{Memories of Passerby I}, the artist examines human memory and identity by blending traditional artistic styles with the unlimited creativity of AI. The artwork demonstrates how AI and human creativity can harmonize, introducing new paradigms into the art world.

In this paper, we align with previous philosophers and artists who advocated for the fluid nature of identity. To explore and express the fluidity of human identity, we employ diffusion models to offer new insights into the parallels between human identity and AI-generated output.

\section{Mechanism of Diffusion Models}

To explore the fluidity of human identity through AI models, we adopt diffusion models, a cutting-edge family of neural networks in AI. Diffusion models are a type of generative AI model that has gained significant attention across various fields, particularly in image generation. These models generate new images from pure random noise through iterative denoising steps. The training procedure of diffusion models can be divided into the following two processes: 1) forward process; 2) denoising (reverse) process.

Specifically, the forward process is the initial phase of training, where noise is progressively added to a given image. This process occurs over many steps (\textit{e.g.}, 1,000), with each step incrementally degrading the image by adding noise. By compiling the outputs of each step, a sequence of images is constructed in which noise is gradually added. This sequence is then utilized in the subsequent denoising process. In the denoising process, a diffusion model is trained to reverse the forward process, learning to progressively denoise the images in the sequence back to their original state. Like the forward process, which consists of many steps, the denoising process also involves multiple steps. The objective is for the model to learn how to recover the image from a noisy state at each denoising step, restoring it to a previous less noisy state.

\begin{figure}[t]
\includegraphics[width=\linewidth]{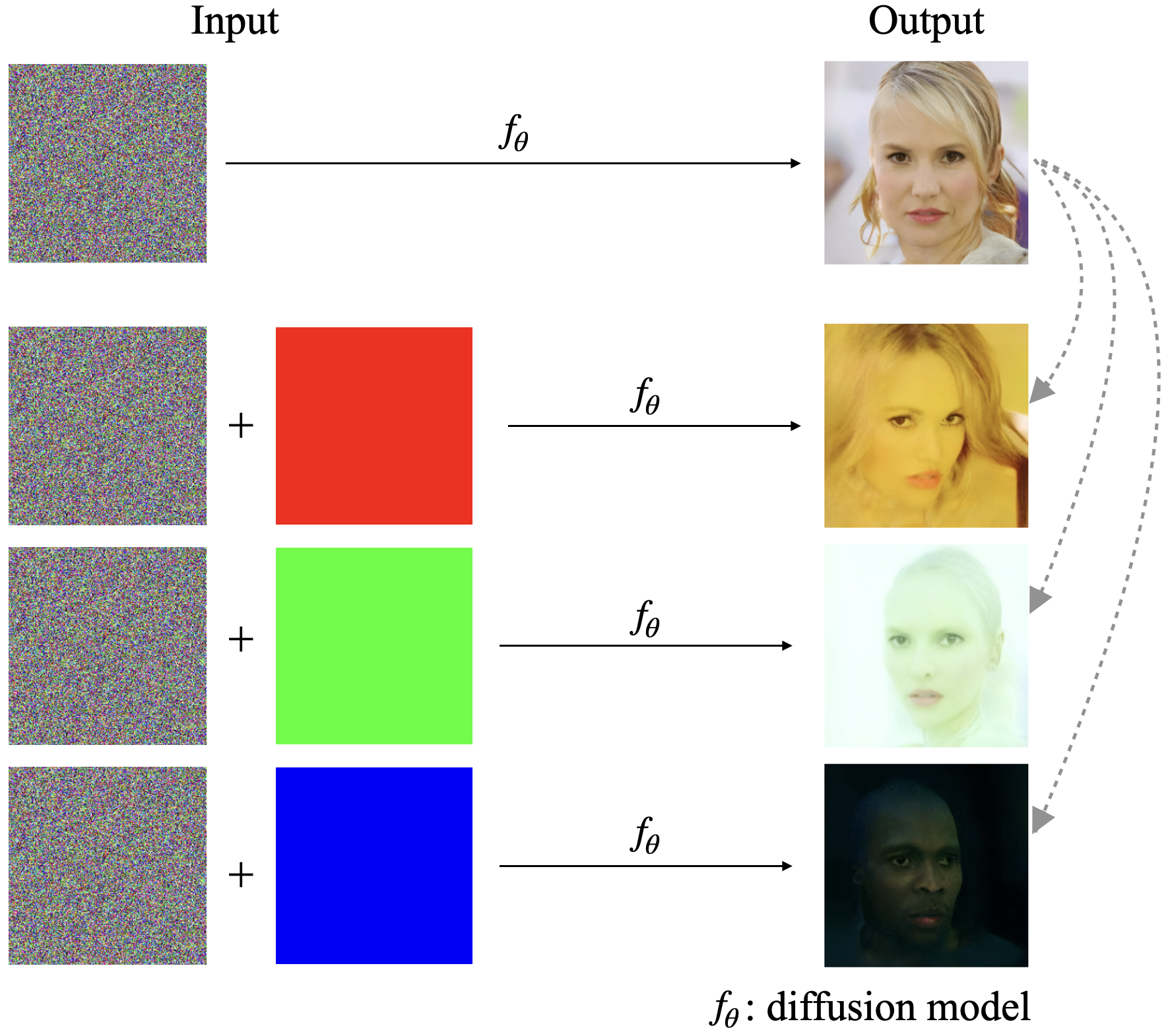}
\caption{Diffusion model outputs based on color-specific additional noise.}
\label{fig:method}
\end{figure}

After the training procedure that consists of these two processes, the model performs the denoising process on pure random noise. Through the iterative steps of denoising process, the model produces new realistic images similar to training images. Depending on the initial noise given, a different image is generated. The overall mechanism of diffusion models are is illustrated in Figure~\ref{fig:mechansim}.

% The diffusion model, a generative AI model, began gaining attention in the field of image generation around 2022~\cite{rombach2022high}. The diffusion model generates images using a principle similar to the thermodynamic diffusion process. The model is trained in two phases: the noising and denoising processes. During the noising process, noise is progressively added over several stages until the image becomes completely random noise. The denoising process, on the other hand, gradually reverses this process by removing the noise step by step, recovering the image. Through this recovery process, the diffusion model learns to reverse random noise back to its original state and also trains a component called the noise predictor~\cite{ho2020denoising}. The noise predictor is designed to generate an output image by reversing noise-added images and making the resulting image similar to the original input. In short, the noise predictor learns to recover the original image by guiding the denoising process. Once trained, the diffusion model can generate meaningful images even when starting with random noise. Due to this mechanism, diffusion models outperform image-generating AI models using GAN technology, especially in terms of image variety and quality.

\begin{figure*}[t]
\includegraphics[width=\linewidth]{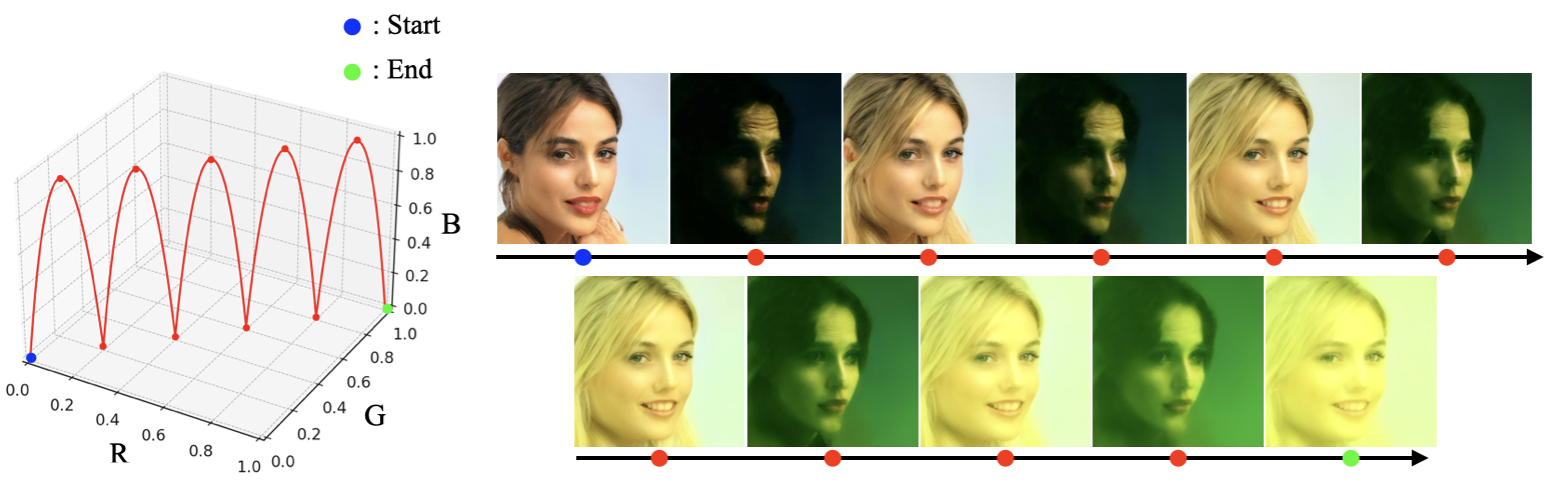}
\caption{Path 1, shown on the left, represents a trajectory resembling a ball bouncing on the ground. The generated human face images on the right illustrate variations as the color of the additional noise changes along this path. Blue and green points indicate the start and end points, respectively, while red points represent sampled color coordinates along the path.}
\label{fig:ex1}
\end{figure*}

\begin{figure*}[h!]
\includegraphics[width=\linewidth]{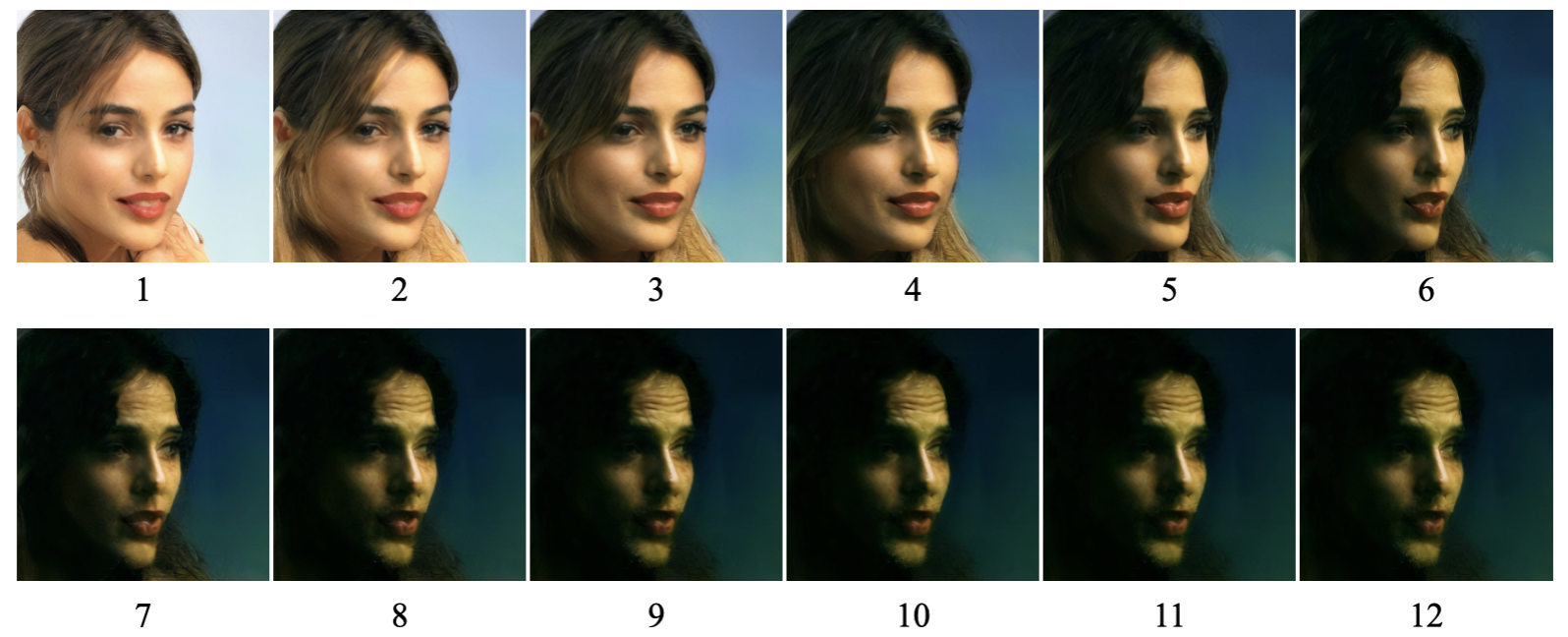}
\caption{Human face images generated when the color of the additional noise is adjusted almost continuously with very small sampling intervals in Path 1.}
\label{fig:sampled}
\end{figure*}

\begin{figure*}[t]
\includegraphics[width=\linewidth]{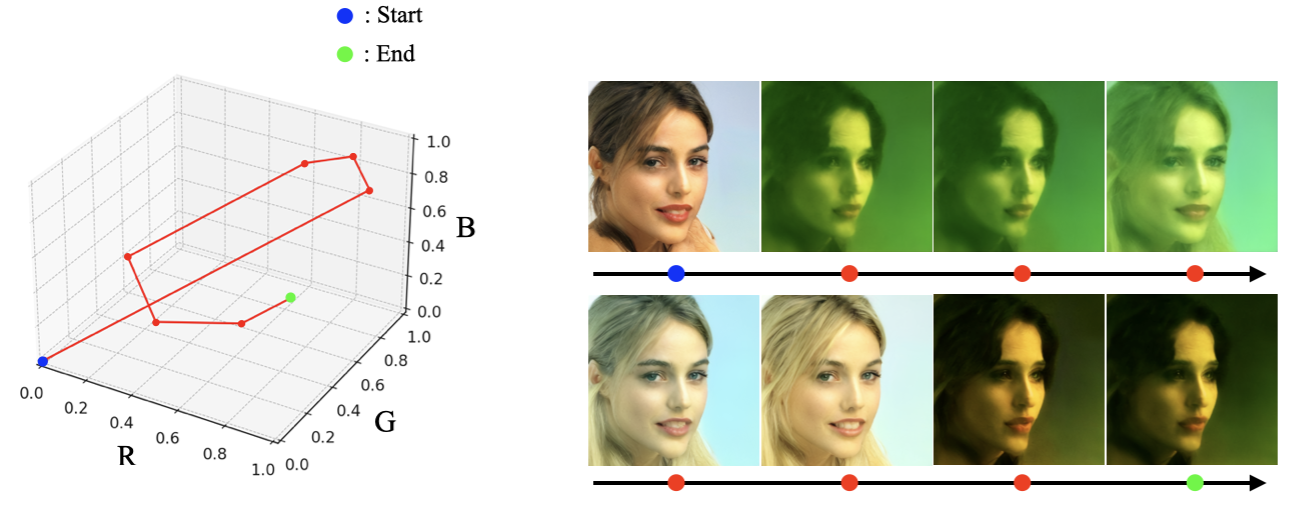}
\caption{Path 2, shown on the left, represents a trajectory resembling light reflecting off mirror surfaces. The generated human face images on the right illustrate variations as the color of the additional noise changes along this path. Blue and green points indicate the start and end points, respectively, while red points represent sampled color coordinates along the path.}
\label{fig:ex2}
\end{figure*}

\begin{figure*}[h!]
\includegraphics[width=\linewidth]{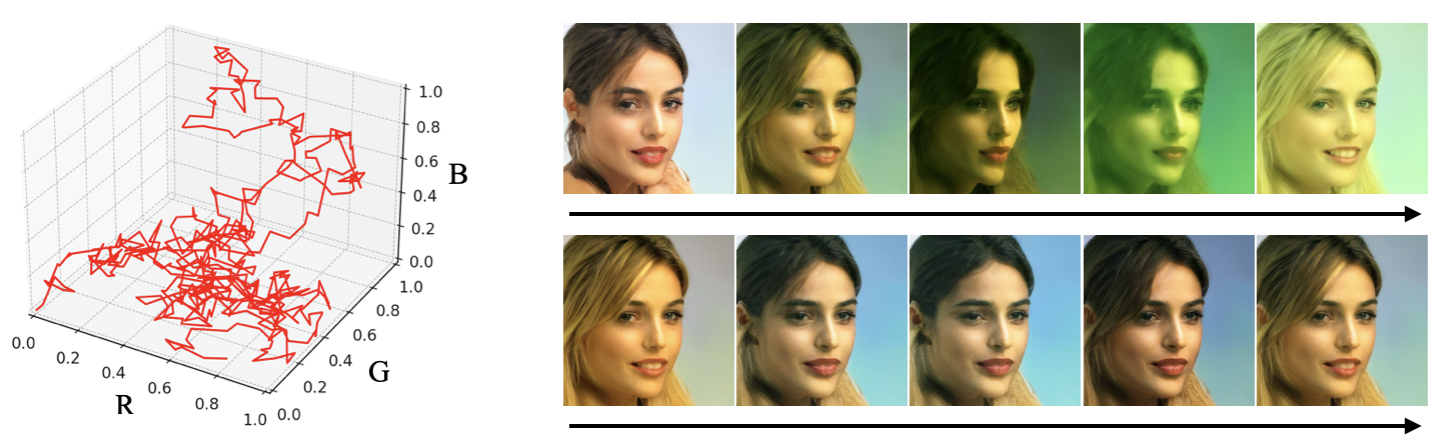}
\caption{Path 3, shown on the left, represents a trajectory similar to Brownian motion. The generated human face images on the right illustrate variations as the color of the additional noise changes along this path. The human face images on the right-hand side display the sampled generated images.}
\label{fig:ex3}
\end{figure*}

\section{Methodology}

Diffusion models generate an image by denoising input noise through a reverse process, which involves a long series of denoising steps.~\cite{yang2023diffusion} 
In this image generation process, we empirically find that injecting additional noise of a specific color into the early stages results in an output image imbued with that color.
In other words, when a diffusion model generates an image by denoising, we can influence the color of the generated image based on the color of the additional noise added to the initial noise.
In addition, we observe that this additional color-specific noise influences not only the color of the generated image but also its content.
Figure~\ref{fig:method} demonstrates the results of a diffusion model when color-specific noise is added to fixed initial noise.
Here, a pre-trained diffusion model generating human face images is employed.
As shown in the figure, compared to the output image without any added noise, the color of the generated image changes according to the color of the additional noise.
We also observe that, in addition to the color, the appearance of the human face in the generated images changes.

This study attempts to explore human identity indirectly, based on the fact that neural networks, including diffusion models, are constructed by mimicking the human nervous system.
Accordingly, we map a diffusion model to the human and the output image of the diffusion model to human identity.
Research in psychology and neuroscience highlights that external factors, such as social interactions and environmental influences, play a significant role in shaping a person’s identity.~\cite{erikson1994identity,oyserman2001self,johnson2001feeling}
We first map initial noise, the input of the diffusion model, to innate characteristics of a person.
Building on the psychological and neuroscientific evidence, we then map additional noise to external factors, where both can be modified.
In the real world, it is impossible to observe a person's identity while exposing them to multiple environments for comparison.
However, with diffusion models generating human faces, we can simulate changes in external input by adjusting the color of the added noise.
To account for the continuous nature of the real world, we gradually vary the color of the additional noise along a specific path in color space.

In summary, our aim is to explore how human identity is influenced by changes in external input. To this end, we observe how variations in additional noise, when added to the initial noise, affect the human facial images generated by the diffusion model. The correspondence can be summarized as follows:
\begin{itemize} \item Image generation process $\leftrightarrow$ human identity formation \item Diffusion model $\leftrightarrow$ human \item Initial noise $\leftrightarrow$ innate factors \item Additional noise $\leftrightarrow$ external factors \item Generated human face image $\leftrightarrow$ human identity \end{itemize}

% To simulate the change in external stimuli, we adjust additional noise by changing the color of the noise.
%

% correspond. bias. human identity.
%

% To convey the idea of investing in the fluidity of human identity by observing generative AI models, we focus on a specific element of image: color. In the process of the reverse process of the diffusion model, we attempt a novel approach named Color Guided Diffusion(CGDiff). CGDiff is a method that can guide the color of output images during the reverse process. For the experiment, we inject noise into the early stage of the reverse steps of image generation and progressively shift the RGB value of color-specific noise. The red-colored path in figure ~\ref{fig:path} is the visualization of the color-specific noise color. The color of the injected noise continuously transforms along a path(figure ~\ref{fig:path}), making dynamic changes to the noise color as it evolves through reverse diffusion.

\begin{figure*}[t]
\includegraphics[width=\linewidth]{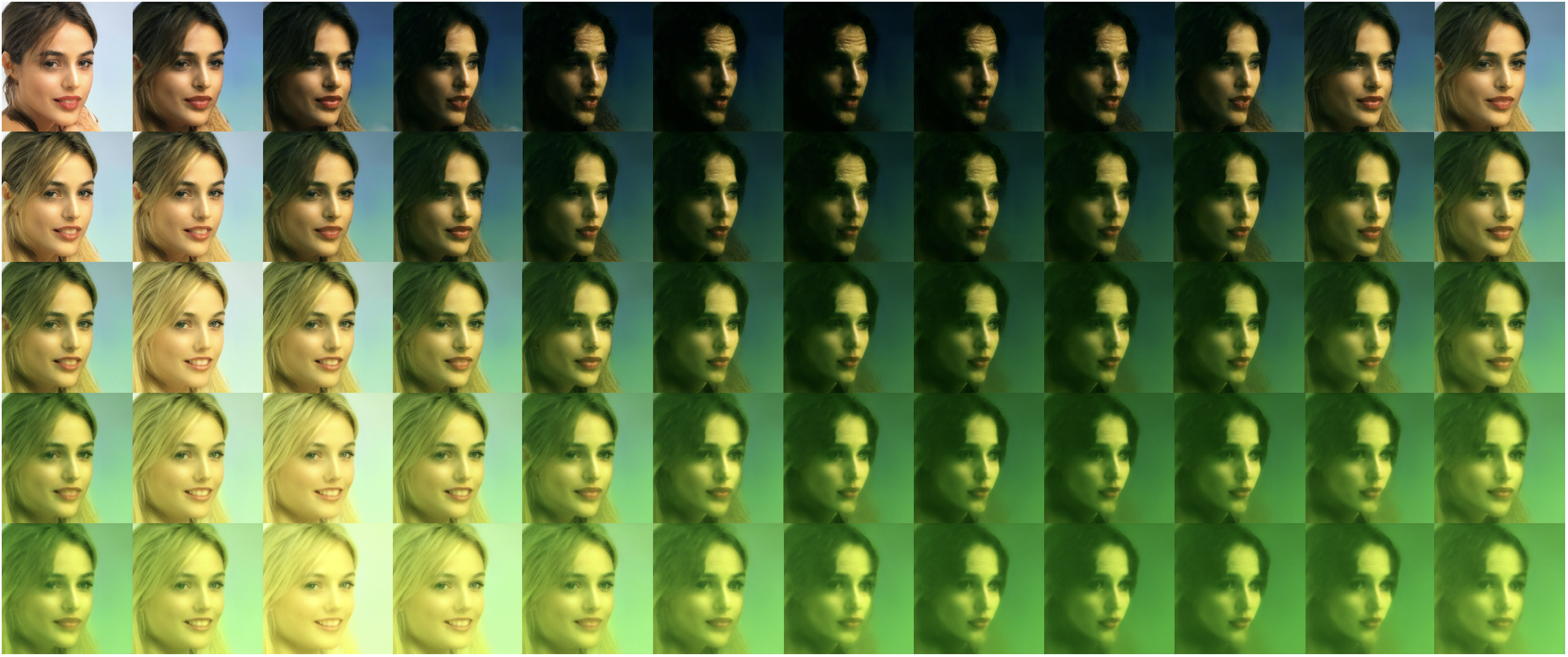}
\caption{Evenly sampled frames of \textit{Fluidity of Human Identity}.}
\label{fig:artwork}
\end{figure*}

\section{Experiments}

In this section, we conduct experiments to analyze the fluidity of human identity based on our experimental finding regarding color-specific additional noise and the results from diffusion models.

\subsection{Experimental Setup}

For experiments, we utilize Google Colaboratory (Colab), which provides free access to GPUs.~\cite{bisong2019google} Through Colab, we run our code on a single NVIDIA Tesla T4 GPU and an Intel(R) Xeon(R) CPU at 2.00GHz. We employ a pre-trained a Latent Diffusion Model (LDM), available in the Huggingface Diffusers library.~\cite{rombach2022high, wolf2019huggingface} This model is trained on the CelebA-HQ dataset, consisting of human face images, enabling it to generate new human face images. ~\cite{karras2018progressive} 

We inject additional noise represented as $s_{noise}\mathbf{M}$, where $s_{noise}$ is a scaling factor and $\mathbf{M}\in \mathbb{R}_{[0,1]}^{3 \times H \times W}$ denotes the color channels of the additional noise. Specifically, $\mathbf{M}$ is normalized so that the R, G, and B values, originally ranging from 0 to 255, are divided by 255 to fit within the [0, 1] range. For instance, if we inject the additional noise of red, the values in the first channel (\textit{i.e.}, the R channel) of M are set to 1, while the values in the other channels are set to 0. We consistently set $s_{noise}$ to 0.01. The additional noise is applied from the first to the tenth denoising step within the denoising process, which consists of a total of 1000 denoising steps.

To analyze the effect of color-specific additional noise, corresponding to external factors, we fix the initial noise representing innate characteristics. We then gradually vary the color of the additional noise, creating three types of paths in the color space as follows:
\begin{itemize} \item Path 1: a trajectory resembling a ball bouncing on the ground \item Path 2: a path simulating light reflecting off mirror surfaces \item Path 3: a trajectory similar to Brownian motion, where a molecule moves randomly without a defined pattern \end{itemize}
Along each path, we alter the color of the additional noise to observe its impact on the generated human face images. For each path, the color of the additional noise is determined by sampling points from the start to the end of the path. Specifically, we input these coordinates in the color space into $\mathbf{M}$ to obtain the generated images corresponding to those points.

% \section{Examination}

% In this section, we review the output images generated from each coordinate along the paths, focusing on how the content and the color of the image evolve concerning the color-specific noise injected during the early stages of the denoising process. We then highlight the correspondence between these results and human nature, further proving AI's legitimacy as a gateway to the human inner self. 

\subsection{Results}

Figure~\ref{fig:ex1} demonstrates the results for the Path 1. On the right-hand side of the figure, the blue and green points represent the start and end of the path, respectively, while the red points represent the main sampled points along the path. The left-hand side of the figure shows the generated human face images corresponding to each color of the additional noise. These results show that as the color of the additional noise changes, the generated human face varies significantly. Notably, the color of each generated face image closely follows the color of the additional noise along the path. Furthermore, the change in not only the color tone but also the shift in appearance and atmosphere is noticeable. Figure~\ref{fig:sampled} shows the generated human face images with minimal sampling intervals. That is, since the color of additional noise is gradually adjusted, the difference in the additional noise between two adjacent images is minimal. We observe that the generated human face images change gradually in response to the variations in additional noise, without any significant alterations. Figure~\ref{fig:ex2} and Figure~\ref{fig:ex3} presents the results for Path 2 and Path 3, respectively. We observe that the diversity in the generated face images depends on the color of the additional noise. Furthermore, the results in Figure~\ref{fig:ex2} and Figure~\ref{fig:ex3} also exhibit the same tendency observed in Figure~\ref{fig:ex1}.

% Figure. illustrates the experiment, demonstrating that the color palette of the image heavily depends on the color-specific noise added at the initial stages of the process. In addition, the change in the content is notable; as the path progresses along different coordinates in the RGB color space, the content goes under transformation. For instance, in all results, the shift in the woman's face based on which color channel predominates the noise at particular path coordinates is noticeable. One crucial observation here is that the change in the shape of the image is not random; it appears to be systematically linked to the change of path and the value of each color channel as the image moves through the noise reduction phase. This also indicates that the transformation of the image is not limited to surface-level shifts like color saturation or brightness but also extends to structural changes in the form and composition of the image. 
% Furthermore, we can notice that the overall mood and tone of the image are also directly related to the color of the noise that is added. For instance, warm-toned noise(red) yields images with intense emotional expression and a brighter mood while cool-toned noise(green, blue) results in a softer, gloomy mood. The change in the mood of the image suggests that color-specific noise plays a crucial role in not only the aesthetic tone of the image but also its structural layers.

\subsection{Fluidity of Human Identity}

Based on our experimental results, we present a video artwork titled \textit{Fluidity of Human Identity}. This video is created by rendering the image sequence generated along Path 1. Figure~\ref{fig:artwork} shows the evenly sampled frames of \textit{Fluidity of Human Identity}. The distinct feature of \textit{Fluidity of Human Identity} lies in its design, which is derived from the observations within the AI model rather than manual creation. Through this artwork, we seek to express the mutable and dynamic nature of human identity, emphasizing that identity is not fixed but subject to change. The strong connection between neural networks and humans inspires viewers to ponder profoundly on the concept of human identity and its relationship with external factors.

% Based on our experimental results, we present a video art- work titled \textit{Fluidity of Human Identity}. The video work demonstrates the algorithmic interplay between dynamically adjusted color-specific noise and a fixed initial state, is central to this artistic contribution. The distinct feature of \textit{Fluidity of Human Identity} lies in its design, which is derived from the observations within the AI model rather than manual creation. Through this artwork, we seek to express the mutable and dynamic nature of human identity, emphasizing that identity is not fixed but subject to change.  The strong connection between neural networks and humans inspires viewers to ponder profoundly on the concept of human identity and its relationship with external factors. Furthermore, the use of AI as a medium to analyze the human psyche challenges the prevailing notion of AI-generated art as mere reproduction and instead advocates for its exploration as a process-driven investigation into identity and perception—core concerns of contemporary digital and generative art. \textit{Fluidity of Human Identity} holds significance in the field of art by positioning AI as a philosophical medium for exploring the formation of human identity as well as bridging artificial intelligence, psychology, and contemporary digital art. 

\subsection{Discussions}

In our experimental results and \textit{Fluidity of Human Identity}, the idea of continuous adjustment along a path offers an analogy for the gradual nature of human identity formation. Similar to the face image generated using both initial noise and additional noise, human identity also goes through transformation depending on innate and external factors. Moreover, since neural networks are structured after humans, this offers a meaningful connection, suggesting that the image generation process portrays the fluid evolution of human identity.

When we expand the analysis to the study of human identity and its development, we can interpret that the nature of human identity fluctuates depending on various external factors and their impact. In this analogy, the color of additional noise, which is selected within a path, could be referred to as the experience, choices, and environment that surround the individuals. Each deviation along the path becomes a unique combination of factors that transforms both the inner and the outer essence of the self. As color-specific noise catalyzes the transformation of the visual output, various external factors may serve as `catalysts' that modify the human identity. Thus, our work serves as a metaphor that supports the concept of fluidity of human identity in response to changes in environmental, emotional, and social contexts. 
The correlation between the color of the additional noise and image generation in diffusion models offers a powerful framework for understanding the characteristics of human identity. This relationship highlights that human identity is a continuously evolving process, influenced by myriad factors, as the path in the color space drives changes in both the appearance and color of the generated face images.

% \textit{Fluidity of Human Identity} holds artistic significance in that it utilizes AI not merely as a means of creation. Our work represents a novel attempt to express human identity based on the correspondence between the human identity formation process and the AI image generation process. This approach suggests future artists

% 
% This work has artistic significance . It suggests future artists to explore it as a process-driven exploration of identity and perception, which is a core concept of contemporary digital and generative art

% TBA TBA TBA

% This work holds significance in the field of art by pushing AI as a philosophical medium to explore the formation of human identity, bridging the fields of artificial intelligence, psychology, and contemporary digital art. The video artwork, \textit{Fluidity of Human Identity}, which is the compliment of the outcome of the algorithmic interplay of dynamically adjusted color-specific noise and fixed initial, is central to the artistic contribution as it visually represents identity as a fluid, evolving concept rather than a fixed entity. Furthermore, the attempt to mobilize AI as a medium to delve into human psyche moves beyond the stereotype of viewing AI art as a mere reproduction. It alsosuggests future researchers to explore it as a process-driven exploration of identity and perception, which is a core concept of contemporary digital and generative art. 

% \subsection{Discussions}
\section{Conclusion}

In this paper, we introduce a novel approach to examine human identity by utilizing generative AI models. Specifically, we employ a diffusion model and analyze the output images it generates depending on the additional noise. Our experiment results demonstrate that the generated images are significantly influenced by the changes in additional noise, revealing the causal relationship between environmental factors and human identity in an indirect manner. Moreover, grounded on these results, we showcase \textit{Fluidity of Human Identity}, which presents the fluid nature of human identity. This study bridges the fields of psychoanalysis, AI, and art, offering valuable insight that AI models can serve as a medium for understanding and expressing humans, given that AI is modeled after human
neurons. We believe that Large Language Models (LLMs), equipped with vast knowledge and capable of mimicking humans, can contribute to exploring the human mind and even be applied to creating innovative artworks.

% a parallel relationship between the impact of noise on the output image and the influence of external factors on human identity. 

% We suggest AI can be a medium for understanding human identity, given that AI is modeled after human neurons.

% Just as color-specific noise alters both the color palette and the content of the image, external factors such as culture and environment have the potential to shift an individual's identity. In summary, this paper demonstrates the fluidity of human identity through AI models and the images they produce.

% \bibliographystyle{chicago}  % Chicago author-date 스타일 적용
% \bibliography{isea} % BibTeX 파일 적용

\bibliographystyle{isea}
\bibliography{iseabib}

\section{Author Biographies}

Yunha Yeo received the B.F.A. degree from the School of the Art Institute of Chicago, Chicago, Illinois, in 2018, and the M.A. degree in Visual Culture Studies from Korea University, Seoul, South Korea, in 2024. She is currently pursuing the M.S. degree at the Graduate School of Cultural Technology, Korea Advanced Institute of Science and Technology, Daejeon, South Korea. Her current research interests include AI-generated art, digital humanities, and media studies.

Daeho Um received the B.S. degree in
electrical engineering from Korea University, Seoul, 
South Korea, in 2019, and the Ph.D. degree in electrical and computer engineering with Seoul National University,
Seoul, South Korea, in 2024. He is currently a Staff
Engineer with Samsung Electronics, Kyeong-Gi, South Korea. His current research interests
include graph-based machine learning, data imputation, and computer vision.

\end{document}